\documentclass[lettersize,journal]{IEEEtran}
\usepackage{amsmath,amsfonts}
\usepackage{algorithmic}
\usepackage{algorithm}
\usepackage{array}
\usepackage[caption=false,font=normalsize,labelfont=sf,textfont=sf]{subfig}
\usepackage{textcomp}
\usepackage{stfloats}
\usepackage{url}
\usepackage{verbatim}
\usepackage{graphicx}
\usepackage{cite}
\usepackage{widetext}
\usepackage{booktabs}
\usepackage{color, colortbl}
\usepackage{pifont}
\usepackage{bm}

\begin{document}

\title{A Safety-Oriented Self-Learning Algorithm for Autonomous Driving: Evolution Starting from a Basic Model}

\author{Shuo Yang, Caojun Wang, Zhenyu Ma, Yanjun Huang$^*$,  Hong Chen,~\IEEEmembership{Fellow,~IEEE}


\thanks{






}


}



\maketitle

\begin{abstract}

Autonomous driving vehicles with self-learning capabilities are expected to evolve in complex environments to improve their ability to cope with different scenarios. However, most self-learning algorithms suffer from low learning efficiency and lacking safety, which limits their applications. This paper proposes a safety-oriented self-learning algorithm for autonomous driving, which focuses on how to achieve evolution from a basic model. Specifically, a basic model based on the transformer encoder is designed to extract and output policy features from a small number of demonstration trajectories. To  improve the learning efficiency, a policy mixed approach is developed. The basic model provides initial values to improve exploration efficiency, and the self-learning algorithm enhances the adaptability and generalization of the model, enabling continuous improvement without external intervention. Finally, an actor approximator based on receding horizon optimization is designed considering the constraints of the environmental input to ensure safety.  The proposed method is verified in a challenging mixed traffic environment with pedestrians and vehicles. Simulation and real-vehicle test results show that the proposed method can safely and efficiently learn appropriate autonomous driving behaviors. Compared reinforcement learning and behavior cloning methods, it can achieve comprehensive improvement in learning efficiency and performance under the premise of ensuring safety.

\end{abstract}


\section{Introduction}
\IEEEPARstart{W}{i}th the continuous development of automotive electronics, sensors and artificial intelligence, autonomous driving technology has become a hot research topic in recent years. This technology is expected to significantly improve transportation efficiency, reduce accidents and lower operating costs, and is the future of transportation systems \cite{feng2023dense}\cite{choi2021horizonwise}. As a key component of autonomous driving technology, the decision-making and planning modules directly affect the level of intelligence of the vehicle \cite{schwarting2018planning} \cite{xu2020adversarial}.

In terms of decision-making and planning methods for autonomous driving, mainstream solutions use rule-based \cite{wang2020research} and optimization-based approaches \cite{chai2018two}. However, the corresponding methods need to manually design a large number of preset rules that are difficult to explore, and the optimization problems involved are also difficult to define and solve, which shows their limitations when dealing with the complex and changing traffic environment \cite{li2023survey}.

In recent years, with the explosive development of artificial intelligence technology, a large number of studies have focused on data-driven decision planning methods \cite{hu2023planning}\cite{deng2021deep}. These methods get rid of the need for complex rule-based design, using a vast quantities of data to facilitate automated algorithmic iteration and optimization. On this basis, self-learning algorithms with experience storage and learning upgrade as the core ideas have garnered increasing attention \cite{silver2017mastering}. The exploration-exploitation mechanism allows such algorithms to learn from new driving data, thereby gradually improving their adaptability to different scenarios. Therefore, it is considered to have the potential to advance the realization of high-level autonomous driving\cite{yang2024guarantee}.

As a representative self-learning method, reinforcement learning (RL) has been employed by numerous researchers to address decision-making and planning problems in autonomous driving \cite{aradi2020survey}\cite{wu2022uncertainty}\cite{li2023ramp}. David et al. proposed an autonomous driving strategy based on deep RL to solve unstop light intersection scenarios \cite{isele2018navigating}. This approach has been demonstrated to outperform common heuristics on several key performance metrics, including task completion time and success rate. He et al. proposed a novel defense-aware robust RL approach with the objective of ensuring the robustness and safety of autonomous vehicles in the face of worst-case attacks \cite{he2024trustworthy}. The experimental results demonstrated that the agent is capable of generating a driving policy that is both reasonable and reliable even when the state input is subject to significant perturbation.

However, the current RL methods still need a lot of exploration and training. This highlights the limitations of low learning efficiency and poor safety, which significantly constrains their application to safety-critical systems, such as autonomous driving\cite{wurman2022outracing}. While some model-based methods mitigate the aforementioned issues to a certain extent, their effectiveness remains significantly affected by the accuracy of the model \cite{moerland2023model}\cite{kaiser2019model}. In addition, some approaches that integrate safety and RL prioritize reducing safety costs, but still have collision rates that fail to achieve comprehensive safety \cite{yang2021wcsac}\cite{junges2016safety}. In particular, these approaches have made progress in reducing risk and enhancing model performance but still face significant challenges when confronted with highly complex and uncertain environments.

This paper proposes a safety-oriented autonomous driving self-learning algorithm to address the above problems. It focuses on the evolution of the given basic model. First, a transformer-encoder-based deep neural network is used to learn a basic model from a small amount of human demonstration trajectories. After that, a policy mixed approach is designed to achieve autonomous algorithmic adaptation and optimization under different scenario inputs. Subsequently, an adjustment mechanism for the discretization of the terminal optimization input is proposed to reduce the optimization search space in a reasonable manner and improve the consistency of the planned trajectories. Finally, an actor approximator based on receding horizon optimization is designed to achieve safe self-learning by constructing constraint functions including static scene and dynamic obstacle. The proposed method is able to perform challenging autonomous driving tasks and achieve good transfer from simulated environments to the real world.

The contributions of this study are summarized as follows:

1) This paper proposes a policy mixed approach that dynamically adjusts the output based on performance feedback, adapting and optimizing the policy to achieve continuous improvement without external intervention.

2) This paper proposes an adjustment mechanism for the discretization of the terminal optimal objective to reduce the optimization search space in a reasonable manner and improve the consistency of the planned trajectories.

3) This paper builds a virtual-real interaction platform based on the autonomous vehicle and high-fidelity simulation software, and deploys the algorithms on the real-world physical vehicle to validate the effectiveness

This paper is organized as follows. The proposed framework is introduced in Section II. The problem formulation is introduced in Section III. The descriptions of self-evolutionary mixed policy method combining basic model are proposed in Section IV. The details of actor approximator based on receding horizon optimization are proposed in Section V. In Section VI, our method is verified and compared in simulation and real-world test, and section VII concludes this paper.

\section{Proposed Framework}\label{sec:overall}

\begin{figure}[!t]
\centering
\begin{tabular}{c}
\includegraphics[width=0.45\textwidth]{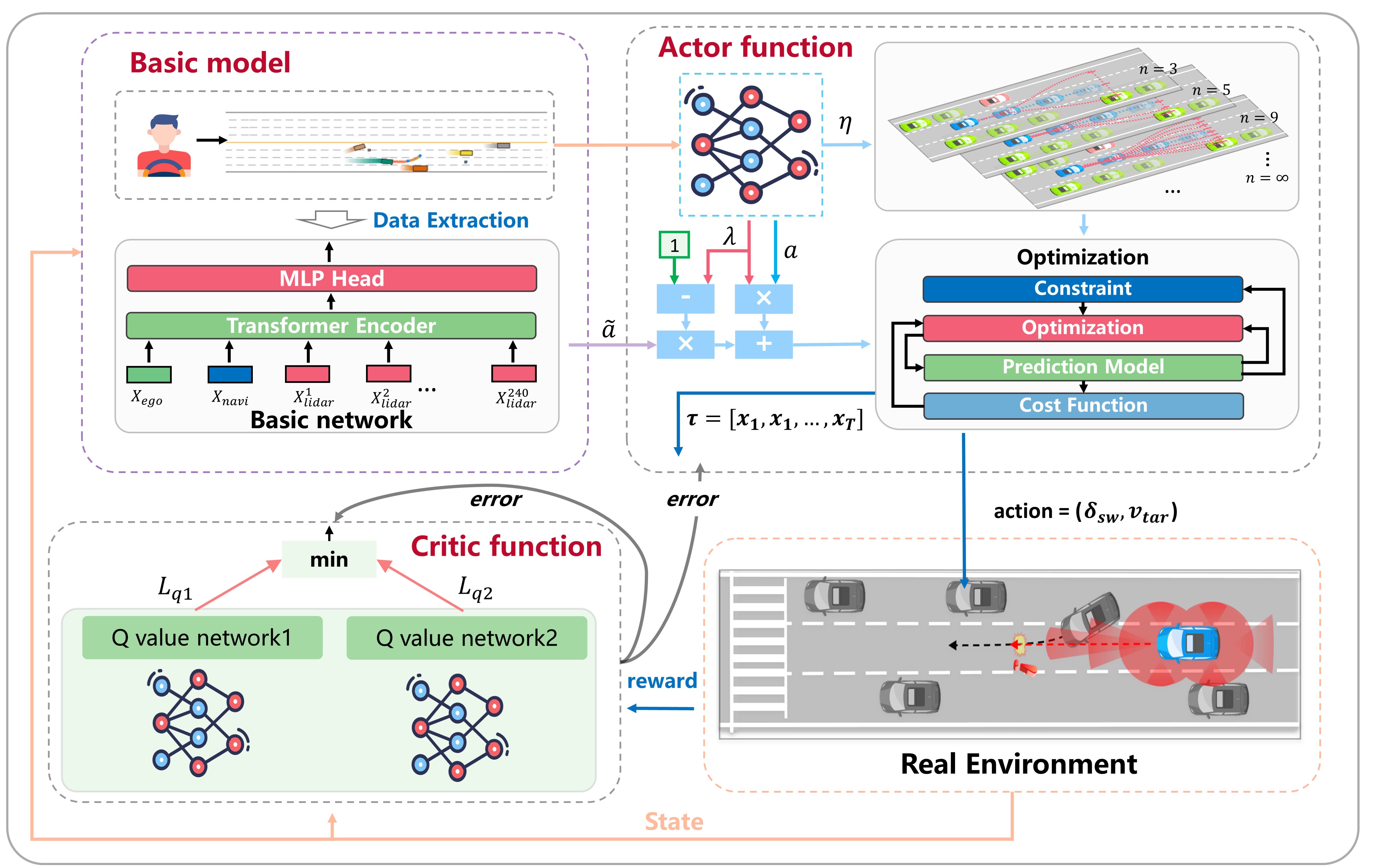}
\end{tabular}
\caption{Overall architecture of safety-oriented self-learning algorithm for autonomous driving.}
\label{overall_architecture}
\end{figure}

The proposed safety-oriented self-learning algorithm for autonomous driving is designed to generate safe and efficient policies. This method not only allows for evolution from a basic model, but also ensures safety in the process of algorithm learning and deployment. The proposed method is based on a typical actor-critic framework to update the policy network and train the value network. Fig. \ref{overall_architecture} depicts the main components of the system, including a basic model based on a transform encoder and an actor function approximator that combines a safety constraint optimization solver and a dynamic policy hybrid module.

The basic model in this framework is trained using demonstration trajectories. The initial policy $\widetilde{a}$ obtained by the basic model is given to the actor function. In the actor function approximator, a fully connected network is used to simultaneously fit the additional policy $\widetilde{a}$, the hybrid proportional parameter $\lambda$, and the discrete factor $\eta$ of the terminal optimization objective. The additional policy  $\widetilde{a}$ is mixed with the initial policy $\widetilde{a}$ and the hybrid degree is dynamically adjusted by the proportional parameter $\lambda$. The discretization factor $\eta$  is used to combine the static scene information to transform it into a discrete dimension n that characterizes the terminal optimal objective. Based on the above output, a receding horizon optimization problem is constructed, combining with the designed constraint function that considers static scene information and dynamic obstacle information, and finally outputs safe and reasonable control actions.



\section{Problem Formulation }\label{sec:Pro_For}

\subsection{Agent Implement}

Markov Decision Process (MDP) is a framework for modeling decision problems. The MDP can be formalized as a quaternion ${(S,A,P,R)}$ , where $S$ is state space, $A$ is action space, $P$ is transition model and $R$ is reward function. Autonomous driving problem can be seen as MDP.

Define $\pi (s)$  as the corresponding action to be taken in a state $s$. The optimal policy can be obtained by solving the Bellman equation with the optimization objective of maximizing the expected cumulative reward.

The action value function ${q_\pi }(s,a)$ as the expected cumulative reward in the future after taking action $a$ from state $s$ can be expressed as:

\begin{equation}
\label{Q function}
\begin{aligned}
{q_\pi }(s,a) &\buildrel\textstyle.\over= {_\pi }\left[ {{G_t}\mid {S_t} = s,{A_t} = a} \right] \\
&={_\pi }\left[ {\mathop \sum \limits_\infty ^{k = 0} {\gamma ^k}{R_{t + k + 1}}\mid {S_t} = s,{A_t} = a} \right],\
\end{aligned}
\end{equation}
where $\gamma$ is discounted factor and ${G_t}$ is return.

The goal of RL is to solve for the optimal policy ${\pi ^*}$ that maximizes the expected cumulative reward. The Soft actor critic (SAC) method is an advanced RL algorithm that has gained widespread use in recent years \cite{haarnoja2018soft}. Its goal is to encourage more extensive exploration by maximizing the entropy objective while maximizing the return, thereby achieving a balance between exploration and utilization. The optimization objectives can be expressed as:

\begin{equation}
\begin{aligned}
&{\pi ^ * } = \arg \mathop {\max }\limits_\pi  {_{\left( {{s_t},{a_t}} \right) \sim {\rho _\pi }}}\left[ {\mathop \sum \limits_t r\left( {{s_t},{a_t}} \right) + \alpha {\cal H}\left( {\pi \left( { \cdot \mid {s_t}} \right)} \right)} \right],\ \\
\end{aligned}
\end{equation}
where ${\cal H}$ is entropy function, $\alpha $ is temperature parameter, with  $\alpha \to 0$, the above optimization objective is gradually approaches that of conventional RL.

\subsection{State Design}

State design is critical to the formulation of RL problems because it directly affects the agent's understanding of the environment. The principles of state design include completeness, abstractness and observability. For autonomous driving, the input information consists of three main parts, i.e. ego state ${S_{ego}}$, navigation state ${S_{navi}}$ and environment state ${S_{lidar}}$ \cite{li2022metadrive}.

The ego state ${S_{ego}}$ is used to provide the agent with information about the ego vehicle. The navigation state ${S_{navi}}$ is used to provide the agent with information about the navigation map. The environment state ${S_{lidar}}$ is the most important for the training of the autonomous driving agent, which is used to represent the information of all the surrounding obstacles. The environment state  is defined as the processed lidar point cloud data. The data starts directly in front of the vehicle and is arranged clockwise.

The above states are normalized. In summary, the state space is defined as:

\begin{equation}
{\rm{s}} = \left[ {{S_{ego}},{S_{navi}},{S_{lidar}}} \right].\
\end{equation}

\subsection{Reward Design}

Reward design represents the fundamental aspect of the agent's interaction with the environment, which defines the agent's objectives. The design of the reward function should be clear and include both positive incentive and negative punishment. For the autonomous driving task, the reward functions are designed as follows:

Autonomous vehicles are encouraged to move along the reference lane towards the destination. Thus, the driving reward ${r_d}$ is defined as maximizing the difference in longitudinal coordinates between two consecutive time steps ${x_{t + 1}} - {x_t}$. It can be expressed as:

\begin{equation}
{r_d} = {c_d} \cdot \left( {{x_{t + 1}} - {x_t}} \right),\
\end{equation}
where ${c_d}$ is driving reward coefficient.

In order to improve the efficiency, the speed reward ${r_v}$ is designed to encourage the agent to travel fast.

\begin{equation}
{r_v} = {c_v} \cdot \frac{{{v_t}}}{{{v_{\max }}}},\
\end{equation}
where ${c_v}$ is speed reward coefficient, ${v_t}$ is current speed, ${v_{max}}$ is maximum speed.

Autonomous vehicles should be safe in all situations, which means that the penalty should be given when any kind of collision occurs. The collision penalty is defined as ${r_c} = {c_c}$, where ${c_c}$ is collision penalty coefficient. And when the agent reaches the target point, it means that the task is completed, and then a sparse success reward ${r_s} = {c_s}$ should be given, where ${c_s}$ is the success reward coefficient.

To sum up, the total reward function is:

\begin{equation}
r = {r_d} + {r_v} + {r_c} + {r_s}\
\end{equation}

The reward coefficients are set as: ${c_d = 1.5}$, ${c_v = 0.5}$, ${c_c = -5.0}$, ${c_s = 10.0}$

\section{Self-evolutionary Mixed Policy Method Combining Basic Model }\label{sec:Self-evolutionary}

\subsection{Mechanism Analysis}

Self-learning algorithms optimize policies through a balance of exploration and exploitation. However, for many tasks, such as autonomous driving, expert demonstration data is readily available. Therefore, the data can be used to quickly learn a basic model through behavioral cloning (BC). And later, a combination with self-learning algorithms is used to allow evolution from a basic model with a certain level of capability.

Based on this approach, some studies have adopted a method of first applying BC for pre-training, followed by RL for fine-tune the policy network. However, these methods fail to consider the inconsistency in policy data distribution between the BC and RL processes, which impacts the optimality and generalization of the algorithm.

In this section, a self-evolutionary mixed policy method combining basic model is proposed. The core idea is as follows: first, use expert data to train a basic model, which outputs an initial policy $\tilde a$. Then, with the weights of the basic model fixed, leverage the self-learning capability of the actor-critic architecture to maximize expected returns by simultaneously optimizing an additional policy $a$ and a mixing proportion parameter $\lambda$. Finally, perform policy mixing based on the output actions (as described in section IV.D).

\begin{figure}[!t]
\centering
\includegraphics[width=2.0in]{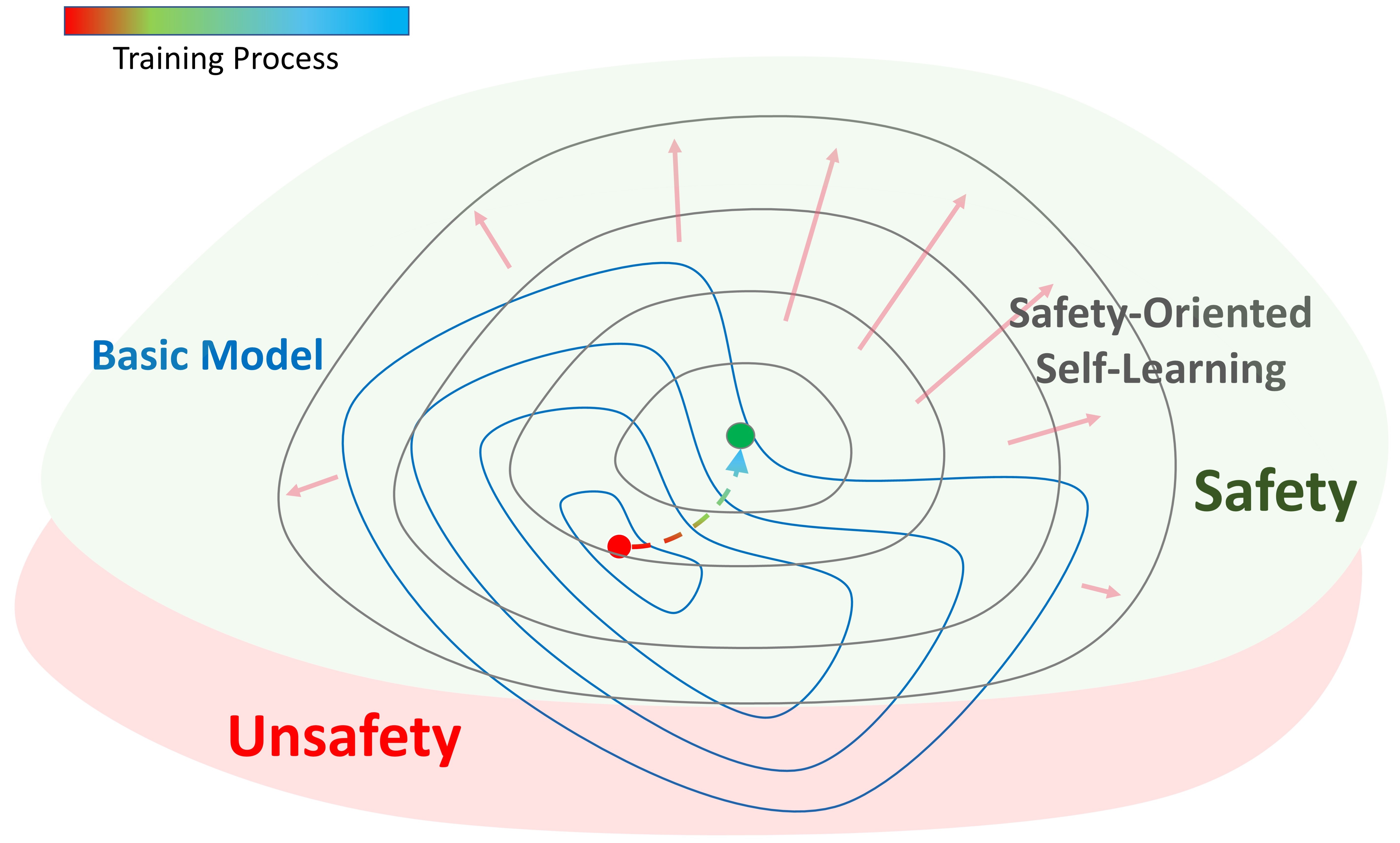}
\caption{The mechanism of self-evolutionary mixed policy method.}
\label{Diagram_of_Mixed_policy}
\end{figure}

The mechanism is illustrated in Fig. \ref{Diagram_of_Mixed_policy}. On one hand, the basic model provides initial values of policy search for the self-learning algorithm, making it easier to explore high-reward areas and thus improving learning efficiency. On the other hand, the combination of the self-learning algorithm and the basic model enables the algorithm to autonomously adapt and optimize across more input scenarios, thereby enhancing model generalization. Furthermore, the design of a safety-oriented actor approximator enables the system to transition from unsafe regions (red) to safe regions (green) during the learning process.

\subsection{Basic Model Training}

Transformer is a deep learning architecture based on the self-attention mechanism, which uses parallelized processing and multi-head attention mechanism to efficiently handle long distance dependencies. Its encoder-decoder structure enables it to perform well in processing sequential tasks and is widely used in the field of autonomous driving.

A Transformer-based basic model is established to map state inputs to control actions. The Transformer encoder consists of two parts: a self-attention encoder and a feed-forward neural network. To improve training stability, residual connections are used to retain information from previous stages' outputs.

The input of the deep neural network is defined in Section III.B. The input vector can be expressed as:

\begin{equation}
\label{STATE}
\begin{aligned}
{\rm{X}} = [ & \underbrace{S_{ego}^1, S_{ego}^2, \ldots, S_{ego}^m}_M, \underbrace{S_{navi}^1, S_{navi}^2, \ldots, S_{navi}^n}_N, \\
                   & \underbrace{S_{lidar}^1, S_{lidar}^2, \ldots, S_{lidar}^o}_O ],
\end{aligned}
\end{equation}
where $M=9$, $N=10$, $O=240$.

The standard Transformer receives as input a 1D sequence of token embeddings. The Transformer encoder model can be expressed as:

\begin{equation}
\label{MSA}
\begin{aligned}
{\bf{z}}_\ell^\prime &= {\mathop{\rm MSA}\nolimits} \left( {\mathop{\rm LN}\nolimits} \left( {\bf{z}}_{\ell - 1} \right) \right) + {\bf{z}}_{\ell - 1}, & \ell = 1...L, \\
{\bf{z}}_\ell &= {\rm MLP} \left( {\mathop{\rm LN}\nolimits} \left( {\bf{z}}_\ell^\prime \right) \right) + {\bf{z}}_\ell^\prime, & \ell = 1...L, \\
{\bf{y}} &= {\rm LN} \left( {\bf{z}}_L \right).
\end{aligned}
\end{equation}

In Eq. \ref{MSA}, the self-attention criterion divides the input embedding into three vectors V, K and Q . The scaled dot-product attention is calculated according to Eq. \ref{QKA}.

\begin{equation}
\label{QKA}
\Theta  = Attention(Q,K,V) = {\mathop{\rm softmax}\nolimits} \left( {\frac{{Q{K^T}}}{{\sqrt {{d_k}} }}} \right)V,\
\end{equation}
where $\Theta$ is scores matrix, $Q$ is a query vector, $K$ is a key vector, $V$ is a value vector, and ${d_k}$ is a normalization.

\begin{figure}[!t]
\centering
\includegraphics[width=3.5in]{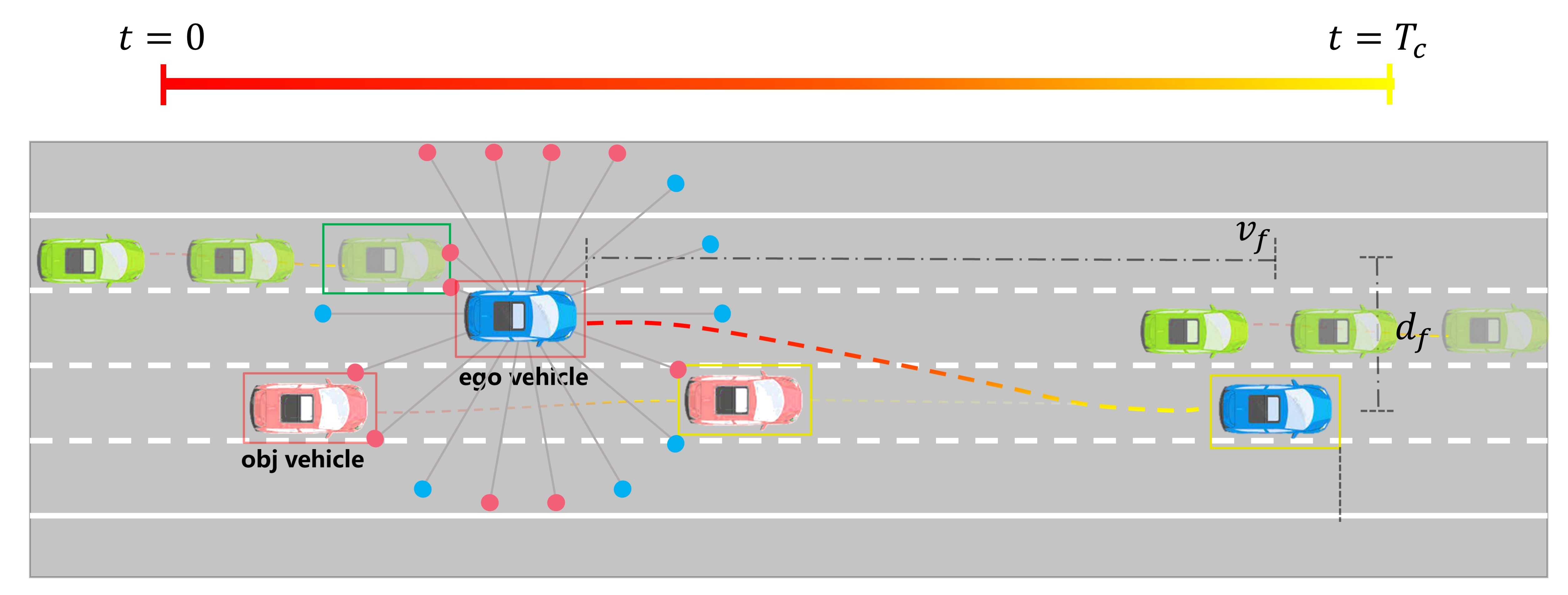}
\caption{The diagram of driving data collection.}
\label{Driving_data_collection_of_DDC}
\end{figure}

Driving data is collected from the human driver using a driving simulator built with a Logitech G29. The recorded data consists of state inputs represented by Eq. \ref{STATE}, as well as the current position and speed. The time stamp for the current position is defined as $0$, and the trajectory is sliced over a period $T_c$. The endpoint coordinates of the trajectory are processed, resulting in the ground truth labels for the basic model outputs as follows:

\begin{equation}
\tilde a = \left[ {{v_f},{d_f}} \right].\
\end{equation}

$\tilde a$ is used to be combined with the additional policy $a$ of the fully connected neural network output in the actor function (In Section IV.C). The mixed policy ${a_{mix}}$ will be input into the receding horizon optimization module (In Section V) to obtain safe action outputs considering the constraints.

\subsection{Self-evolutionary mixed policy method}

The policy mixing process can be expressed as follows:

\begin{equation}
{a_{mix}} = \tilde a\left( {1 - \lambda } \right) + a\lambda \
\end{equation}

In the training process, the fully connected neural network in the actor function simultaneously learns $\lambda$ and $a$. As described in Section II.A, the optimization objective is to maximize the average return. As the model continues to converge, the learning process can be expressed as follows:

a) When $\lambda  \to 0$, it indicates that the action $\tilde a$ output by the basic model yields a higher average return. This means that in these states, the basic model is already performing well, so the role of the RL model becomes relatively less important. Meanwhile, the basic model also guides the RL model to explore more high-reward state-action pairs, which improves the learning efficiency of the RL model.

b) When $\lambda  \to 1$, it indicates that the reward corresponding to the action $\tilde a$ output by the basic model is low, which significantly affects the performance of the algorithm. In this case, the RL model needs to play a more significant role, adjusting both the parameter $\lambda$ and the additional policy $a$ to maximize the average return. This also means that in these scenarios, the generalization ability of the basic model is insufficient, and the RL model must compensate for the policy to ensure the performance.

Through the design of the aforementioned mixed policy method, the roles of the basic model and the RL model can be effectively balanced. The RL model dynamically compensates for the output of the basic model based on performance feedback in different scenarios, thereby optimizing the overall policy. This method enables autonomous adaptation and optimization, promoting continuous improvement without external intervention.

\subsection{Adjustment Mmechanism for the Discretization of the Terminal Optimal Objective}

The output of the mixed policy ${a_{mix}}$ is defined in Section IV. B and C. The first and second dimensions of ${a_{mix}}$ represent the expected longitudinal velocity ${v_f}$ and expected lateral offset ${d_f}$ after time $T_c$, respectively. These outputs are passed to the downstream receding horizon optimization to generate optimized trajectories and control actions that satisfy safety constraints.

From the perspective of policy search, continuous action outputs are generally better at considering optimality compared to discrete actions. However, for autonomous driving tasks, the following factors still need to be considered:

a) In some tasks, it is more consistent with human habits to perform as few actions as possible. This is because human drivers usually choose a more stable and consistent control method in simple scenarios. Therefore, the algorithm design of autonomous driving should also mimic this behavior to achieve more natural and efficient driving. In such cases, discrete actions are appropriate.

b) The consistency of the planned trajectory must be considered. This is because inconsistent trajectories can lead to vehicle instability and passenger discomfort. If the trajectory of the vehicle changes frequently, it may also pose challenges to downstream motion control and increase the risk of accidents. Therefore, considering discrete action outputs can effectively mitigate this problem.

To solve the above trade-off problem, this paper proposes the concept of discretization degree and quantifies it by the design of the discretization factor $\eta$. The discretization of lateral planning trajectories is concerned in this paper.

Specifically, RL outputs $\eta$ to automatically search for the appropriate level of discretization based on the complexity of the task. The discreteness factor $\eta$ is used in conjunction with static scene information to convert it into a discrete layer number $n$ that represents the granularity of optimization at different endpoints.

First, the discretization factor $\eta$ is reversely normalized as:

\begin{equation}
n = \left\lfloor {\kappa  \cdot \eta  + \sigma } \right\rfloor, \
\end{equation}
where $\kappa$ is proportionality coefficient and $\sigma$ is offset coefficient.

Define ${l_{\max }}$ and ${l_{\min }}$ as the distance from the ego vehicle to the left and right boundaries of the road, respectively.

\begin{equation}
\begin{aligned}
    &\Delta x = \frac{{{l_{\max }} - {l_{\min }}}}{n},\ \\
    &{P_i} = {l_{\min }} + i \cdot \Delta x,\begin{array}{*{20}{c}} \
    {}&{i = 0,1,...,n,}
    \end{array}\ \\
    &\delta {f_{closest}} = \mathop {\min }\limits_{{P_i}} \left| {{P_i} - {d_f}} \right|\
\end{aligned}
\end{equation}
where $\Delta x$ is discrete interval. The discrete point $\delta {f_{closest}}$ closest to ${d_f}$ is calculated as follows:

It can be seen that when $\eta  \to \infty$, the parameter search of the lateral trajectory planning is performed in continuous space. However, when $\eta  \to 1$, the lateral planning degenerates into following the global path. In this paper, $\eta $ is automatically tuned through RL to output the most appropriate $\eta $ in different scenarios.

Finally, the mixed policy ${a_{mix}}$ is defined as ${a_{mix}} = \left[ {{v_f},\delta {f_{closest}}} \right]$.

\section{Actor Function Approximator Based on Receding Horizon Optimization}\label{sec:Actor function approximator}

\subsection{Overall Architecture of Actor Approximator}

The actor approximator is used to interact directly with the environment and learn optimal policies through optimization iterations. The actor approximator consists of three main components: a deep neural network, a receding horizon optimizer, and a policy mixed component. The deep neural network is used to model the mapping between the environment state and the optimal action, and the performance of the algorithm is continuously improved through gradient updates. The policy mixed component is used to combine the policies generated by RL and BC in order to merge the advantages of both (In Section IV).

In this section, the construction of the receding horizon optimization(RHO) is presented. This component is used in combination with two other components to output a safe and reasonable vehicle control value.

\subsection{Optimization Problem Design}

\subsubsection{Kinematic Vehicle Model}

In this paper, the kinematic model is used to represent the vehicle motion. The kinematic bicycle model is defined by the following set of differential equation:

\begin{equation}
\begin{array}{l}
x = v \cdot \cos \varphi \\
y = v \cdot \sin \varphi \\
\varphi  = v \cdot \frac{{\tan ({\delta _f})}}{L},
\end{array}\
\end{equation}
where ${\delta _f}$ is front wheel angle, $L$ is wheel base.

The discrete-time form of model at time step $i$ is:

\begin{equation}
\label{state_transfer}
{\zeta _{i + 1}} = f({\zeta _i},{u_i}),\
\end{equation}
where ${\zeta _i} = {\left[ {\begin{array}{*{20}{c}}{{x_i}}&{{y_i}}&{{\varphi _i}}\end{array}} \right]^T}$ is the state, and ${u_i} = {\left[ {\begin{array}{*{20}{c}}
v&{{\delta _f}}\end{array}} \right]^T}$ is the input. The Eq. \ref{state_transfer} is calculated with the sampling interval $T$.

\subsubsection{Cost function design}

Autonomous driving systems are required to be able to reach the destinations safely and comfortably with the highest possible efficiency. The nonlinear programming problem is constructed, and the safety actions can be obtained by solving constrained optimization problems.

The optimization problem is specified as:

\begin{equation}
\begin{aligned}
{J_{\pi MPC}} = &{\rm{ }}\sum\limits_{j = 1}^m {\left\| {{\zeta _{t + {N_p}\mid t}} - \zeta _{des,t + {N_p}\mid t}^j} \right\|_{{Q_j}}^2}  + \sum\limits_{i = t}^{t + {N_c}} {\left\| {{u_{i\mid t}}} \right\|_{{R_u}}^2} \\
                &+ \sum\limits_{i = t}^{t + {N_c}} {\left\| {\Delta {u_{i\mid t}}} \right\|_{{R_{du}}}^2} \
\end{aligned}
\end{equation}
where $N_p$ is control horizon, $N_c$ is prediction horizon, $Q_j$, $R_u$ and $R_{du}$ denote the weight coefficients of each cost function.

The terminal states set is expressed as $\zeta _{des,t + {N_p}|t}^j = {\left[ {\begin{array}{*{20}{c}}{x_{des,t + {N_p}|t}^j}&{y_{des,t + {N_p}|t}^j}&{\varphi _{des,t + {N_p}|t}^j}\end{array}} \right]^T}$. $\zeta _{des,t + {N_p}\mid t}^j$ is calculated from the output ${a_{mix}} = \left[ {{v_f},\delta {f_{closest}}} \right]$ of the policy mixed component. Specifically, taking the current vehicle state $\zeta $, and the desired velocity ${v_f}$ and the desired lateral distance $\delta {f_{closest}}$ after the ${N_p}$ steps as inputs, the transition from the current state to the target state is fitted by a fifth-degree polynomial, and the polynomial coefficients are solved by the equations system satisfying  the conditions of the initial and target positions, velocities, and accelerations to obtain a smoothed trajectory ${f_{traj}} = \left[ {\zeta _{poly}^1,\zeta _{poly}^2,...,\zeta _{poly}^{{N_p}}} \right]$.

That is,$\zeta _{des,t + {N_p}|t}^j = {\left[ {\begin{array}{*{20}{c}}{\zeta {{_{poly}^{{N_p}}}_{[0]}}}&{\zeta {{_{poly}^{{N_p}}}_{[1]}}}&{\zeta {{_{poly}^{{N_p}}}_{[2]}}}\end{array}} \right]^T}$

\subsubsection{Constraint design}

There are three constraint terms, including vehicle kinematics constraint, change rate of control input constraint and obstacle avoidance constraint, which are defined as follows:

\begin{equation}
\begin{aligned}
            &\xi_t = f(\xi_{t-1}, a_t) \quad \text{ } t = 1, 2, \ldots, N_C - 1, \\
            &\Delta a_{\text{min}} < \Delta a_t < \Delta a_{\text{max}} \quad \text{} t = 1, 2, \ldots, N_p, \\
            &{y_{t + {N_p}\mid t}} + {\rm{\Delta }}x \cdot {\varphi _{t + {N_p}\mid t}} - w/2 > {y_{{\rm{obs}},r}} + {w_{{\rm{obs}}}}/2,\ \\
            &{y_{t + {N_p}\mid t}} + {\rm{\Delta }}x \cdot {\varphi _{t + {N_p}\mid t}} + w/2 < {y_{{\rm{obs}},l}} - {w_{{\rm{obs}}}}/2,\ \\
            &\forall \left( {{x_{fr,t + {N_p}\mid t}} > {x_{{\rm{obs}}}}} \right) \cap \left( {{x_{rr,t + {N_p}\mid t}} < {x_{{\rm{obs}}}} + {l_{{\rm{obs}}}}} \right),\
\end{aligned}
\end{equation}
where $\Delta {a_{\min }}$ and $\Delta {a_{\max }}$  are upper and lower bounds of control input change rate, respectively. $w$ is the width of the ego vehicle, ${w_{{\rm{obs}}}}$ is the width of the obstacle vehicle, ${l_{{\rm{obs}}}}$ is the length of the obstacle vehicle. ${y_{{\rm{obs}},l}}$ and ${y_{{\rm{obs}},r}}$ are the maximum and minimum values of the lateral coordinates of the four corner points of the obstacle vehicle, respectively. ${x_{fr,t + {N_p}\mid t}}$ and ${x_{rr,t + {N_p}\mid t}}$ are the maximum and minimum values of the longitudinal coordinates of the four corner points of the ego vehicle, respectively. And ${x_{{\rm{obs}}}}$ is the longitudinal coordinate of the obstacle vehicle.

\section{Experiments Verification}

\begin{figure}[!t]
\centering
\includegraphics[width=3.5in]{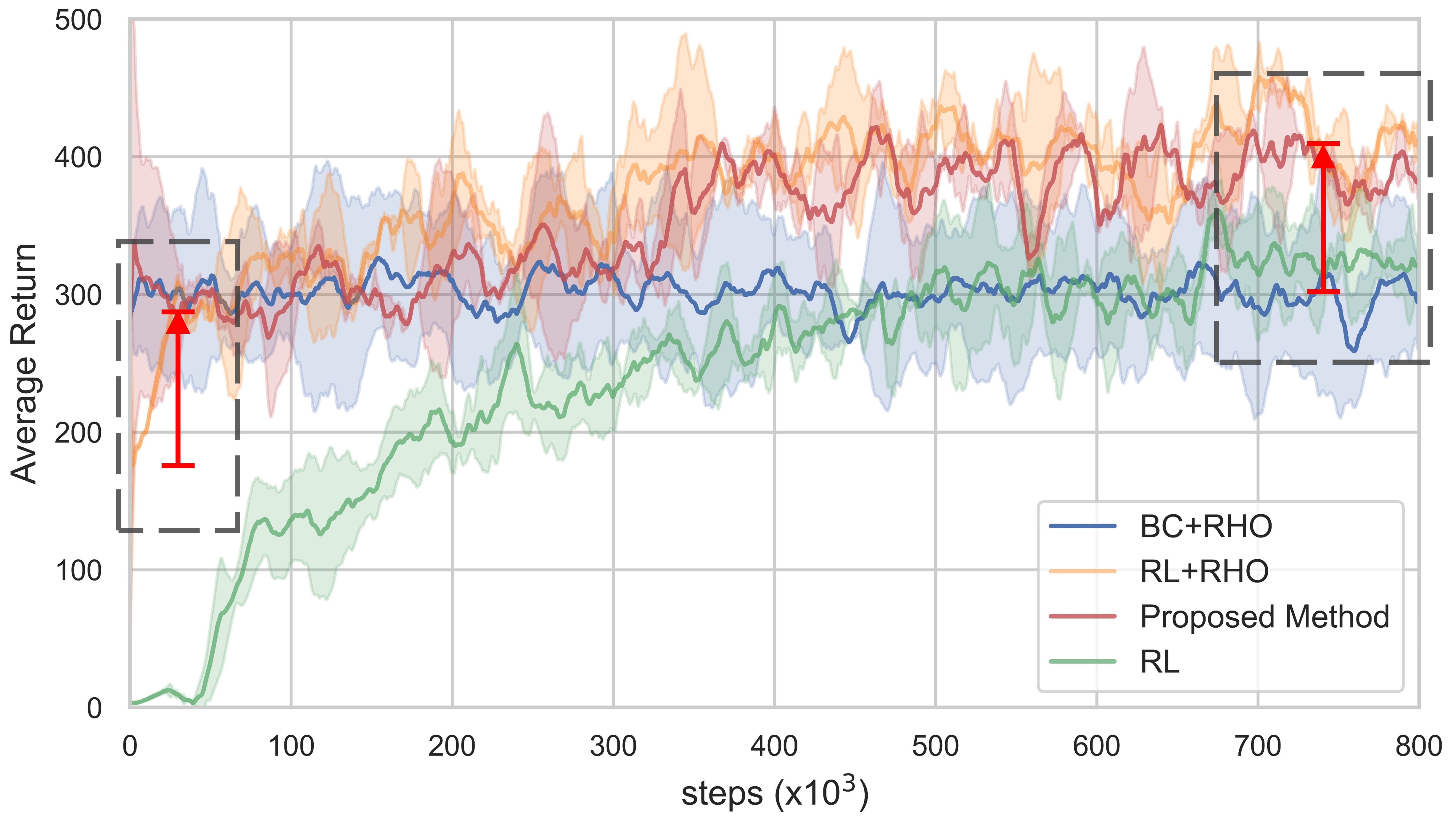}
\caption{Comparison of performance among the proposed method and the baselines}
\label{average return}
\end{figure}

\begin{table*}[!t]
    \centering
    \begin{tabular}{ccccc}
        \toprule
        \textbf{Stats} & \textbf{RL} & \textbf{Rule Based} & \textbf{BC+RHO} & \textbf{Proposed Method} \\
        \midrule
        Cumulative reward & 343.67 & 378.91 & 337.78 & \textbf{415.04} \\
        Avg. Speed & \textbf{12.94} $\pm$ 2.95 $m/s$ & 7.07 $\pm$ 0.98 $m/s$ & 9.28 $\pm$ 0.74 $m/s$ &9.33 $\pm$ \textbf{0.48} $m/s$ \\
        Collision Rate & 10\% & 3.3\% & 6.6\% & \textbf{0\%} \\
        \bottomrule
    \end{tabular}
    \caption{Comprehensive performance comparison of RL, Rule based, BC+RHO, and the proposed method after 30 runs.}
    \label{tab:comparison}
\end{table*}

In this section, the proposed method is validated in a mixed traffic scenario that includes pedestrian crossings with random initial locations and speeds. The traffic vehicles use the IDM and MOBIL models. The training environment was built using the simulation software MetaDrive \cite{li2022metadrive}. The optimization problem solver selects IPOPT \cite{biegler2009large}.

\subsection{Simulation results}

\begin{figure}[!t]
\centering
\includegraphics[width=3.5in]{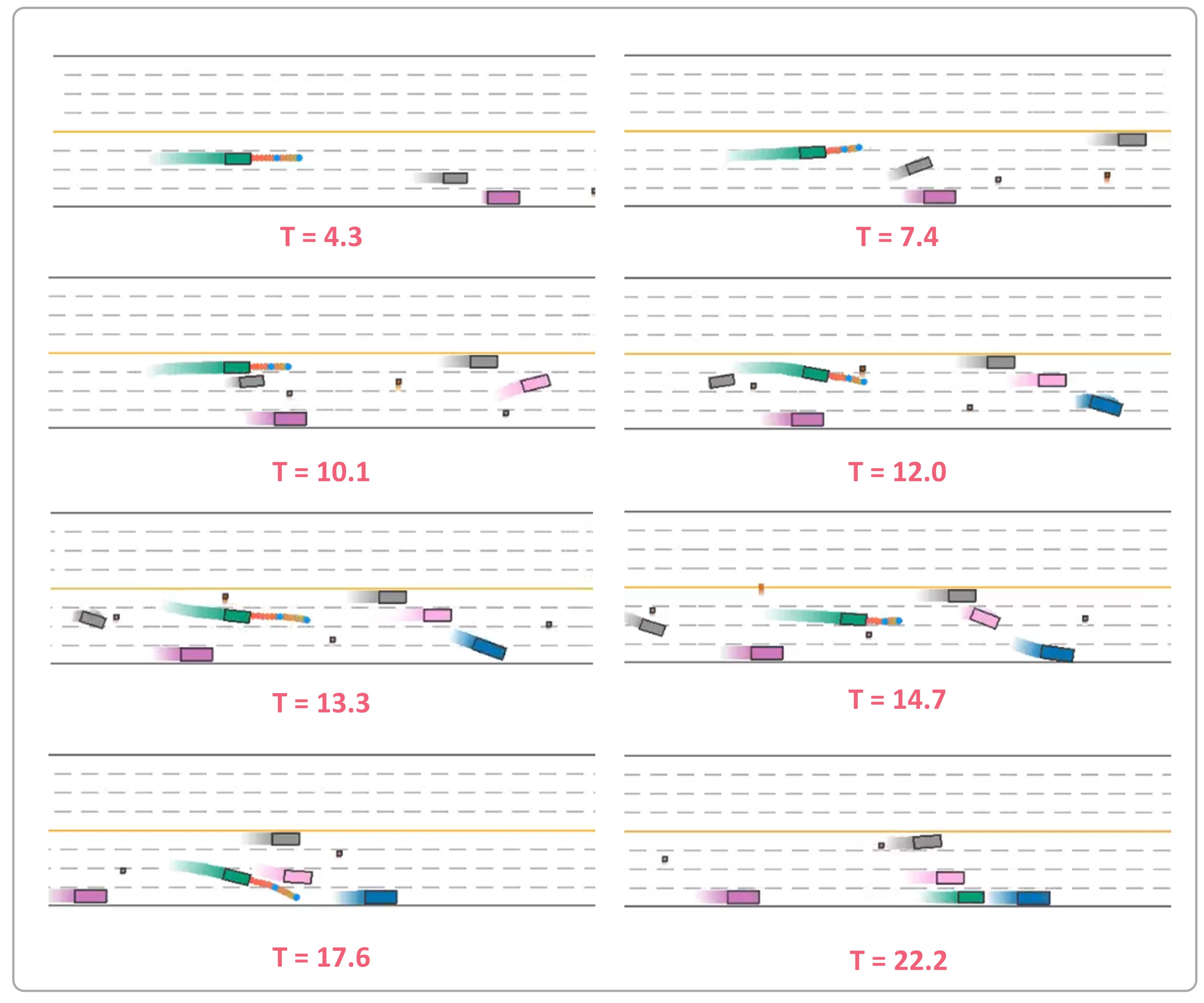}
\caption{Response results of the proposed algorithm for mixed dynamic traffic flow}
\label{sim_vehicle_traj}
\end{figure}

The algorithm is deployed in the training environment. The computer is equipped with an Intel core i7-13700F CPU, NVIDIA GeForce GTX 4070Ti GPU.

The ablation experiments are conducted to demonstrate the effectiveness of the proposed components. In this paper, BC+RHO, RL+RHO, and an RL algorithm that directly outputs control actions are used as baselines. The comparison of performance between the proposed method and the baselines is shown in Fig. \ref{average return}. It can be seen that after 80,0000 training steps, the average return of the proposed algorithm steadily increases, surpassing all baseline algorithms. Additionally, the proposed method avoids the early-stage low performance seen in RL+RHO, achieving evolution from the basic model and leading to comprehensive improvements in learning efficiency and performance metrics.

Fig. \ref{sim_vehicle_traj} shows the response results of the proposed algorithm for mixed dynamic traffic flow. It can be seen that the proposed method, even in highly challenging scenarios with multiple pedestrians crossing randomly (a challenge even for human drivers), can safely and quickly respond to changes in traffic flow through reasonable acceleration, deceleration, lane changes, and overtaking maneuvers. TABLE. \ref{tab:comparison} shows the comprehensive performance comparison of RL, rule-based(IDM+MOBIL), BC+RHO, and the proposed method after 30 runs. It can be seen that the proposed method has significant advantages in terms of cumulative reward and collision rate. Although the average speed is lower than that of RL, the higher speed of RL comes at the expense of safety.

\subsection{Real Vehicle Test Results}

\begin{figure}[!t]
\centering
\includegraphics[width=3.5in]{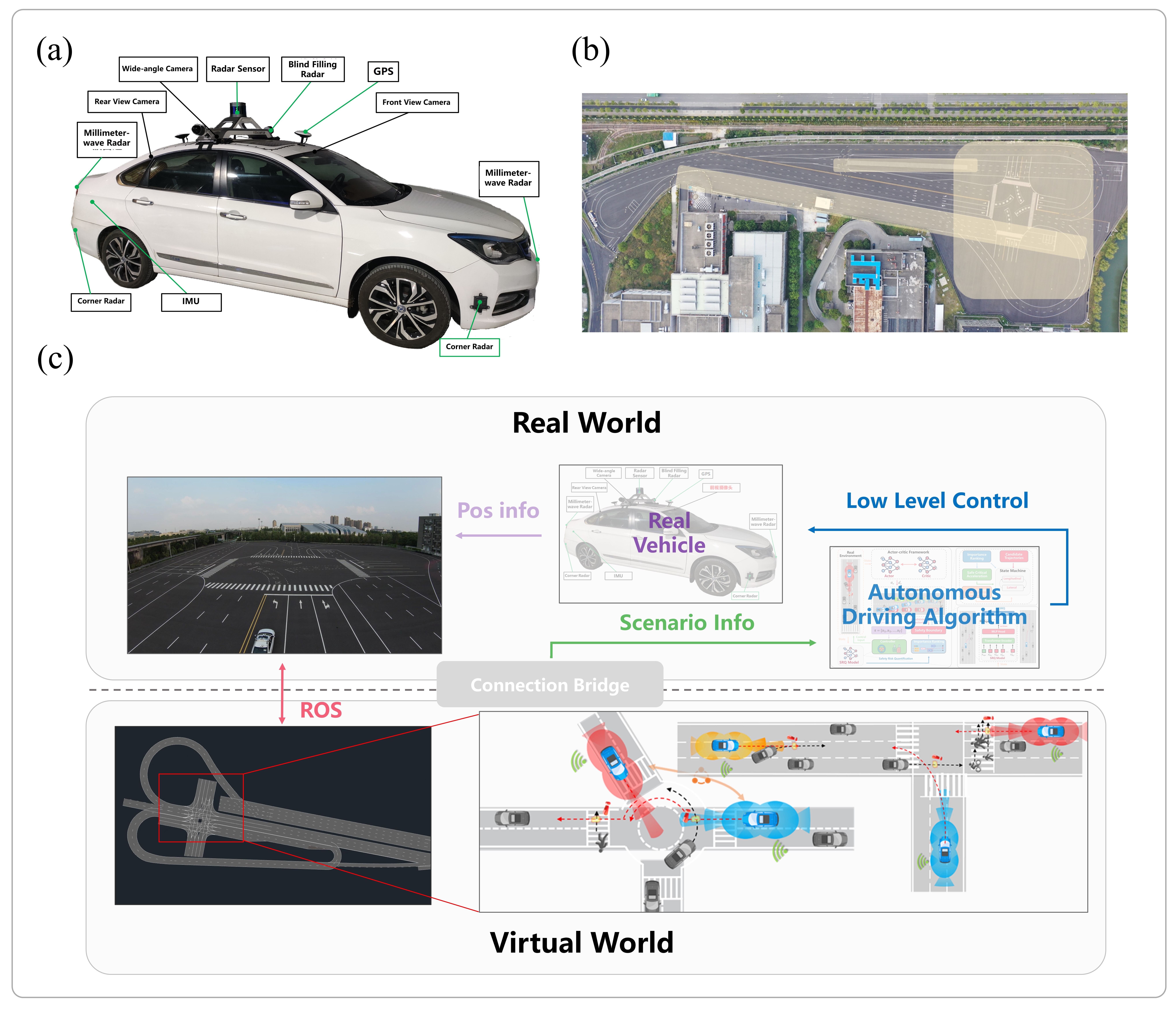}
\caption{The overall architecture of the virtual-real interaction platform. a) Hardware design of autonomous vehicle. b) The test filed of Intelligent vehicles in Tongji university, Shanghai. c) Vitural-real interaction platform}
\label{virtual-real}
\end{figure}

In order to further illustrate the effectiveness of the proposed algorithm, the real vehicle test is carried out. The algorithm is deployed on an autonomous vehicle equipped with rich sensors and high-performance computers, as shown in Fig. \ref{virtual-real}(a). The hardware system of the vehicle includes integrated navigation system, Lidar, millimeter-wave radar, camera, power supply, industrial computer, etc. The CPU of the industrial computer is Intel i9-9900k and the GPU is RTX-3080Ti, including Ethernet and CAN interfaces.

As shown in Fig. \ref{virtual-real}(c), we use high-fidelity simulation software and real autonomous vehicle to build this self-evolution platform based on virtual-real interaction, and realize the preliminary verification of the self-evolution algorithm on this platform. Specifically, autonomous vehicles run in both the real and virtual world, dealing with both real and virtual traffic scenarios. The software architecture is based on ROS, a distributed software framework for robot development. The experimental site is selected in the closed test field of Tongji University in Shanghai, which includes 430 $ m$ standardized test road sections, which can meet the test requirements of the algorithm, as shown in Fig. \ref{virtual-real}(b).

\begin{figure}[!t]
\centering
\includegraphics[width=3.5in]{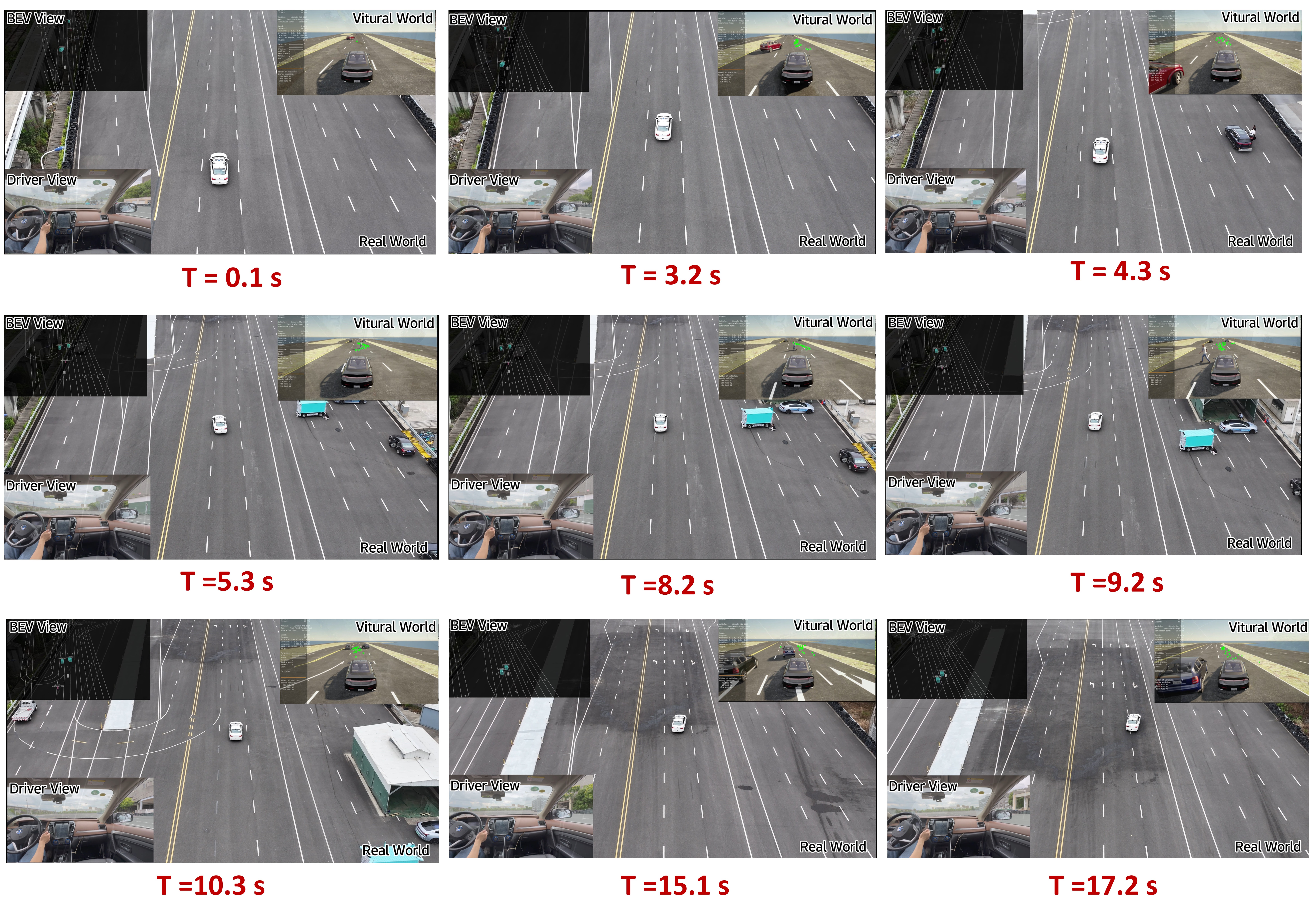}
\caption{Experimental results of real vehicle test under bird's eye view}
\label{real_vehicle_BEV}
\end{figure}

\begin{figure}[!t]
\centering
\includegraphics[width=3.5in]{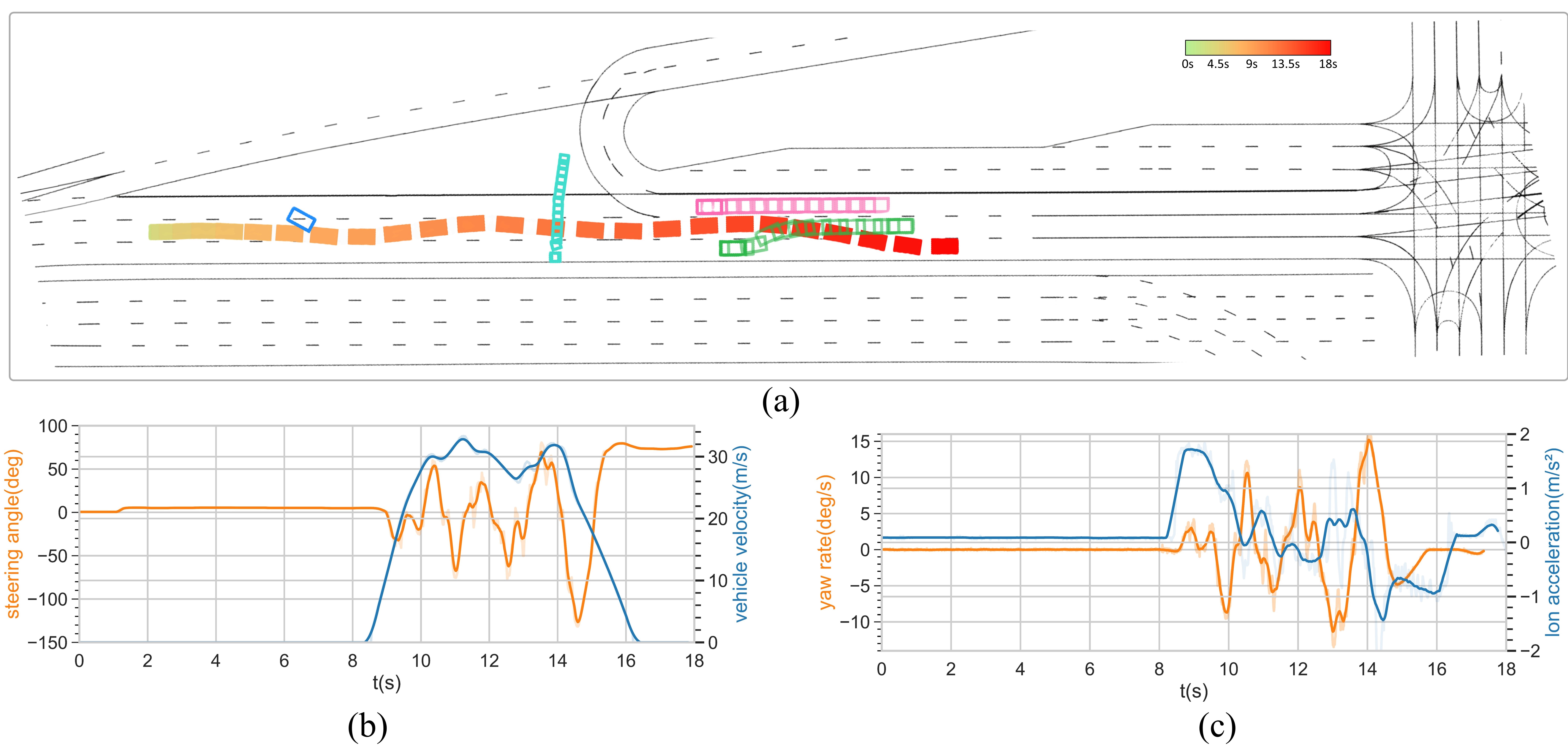}
\caption{Sensor recording results. a) Heading vs Timesteps, b) steering angle vs Timesteps, c) vehicle velocity vs Timesteps, d) yaw rate vs Timesteps, e) longitudinal acceleration vs Timesteps, f) lateral acceleration vs Timesteps, respectively.}
\label{Sensor_real_results}
\end{figure}

The bird's-eye view results of the real vehicle experiments are presented in Fig. \ref{real_vehicle_BEV}. The designed scenario encompasses continuous events, including a stationary obstacle diagonally intruding into the current lane, a suddenly crossing pedestrian, and a dynamic vehicle cutting into the current lane. The results illustrate that the proposed algorithm effectively navigates around static obstacles, flexibly avoids crossing pedestrians, and appropriately responds to suddenly appearing traffic vehicles. This indicates that the proposed method is well-suited for application in real-world autonomous driving systems. The sensor recording results are depicted in Fig. \ref{Sensor_real_results}.

\section{Conclusion}

This paper proposes a safety-oriented self-learning algorithm to enhance the self-evolution capability of autonomous driving in complex environments. By designing a basic model based on a transformer encoder and a dynamic mixed policy method, the learning efficiency and policy optimization ability are improved. Additionally, an actor approximator based on RHO ensures the system's safety. Simulation and real-vehicle test results show that the proposed method can safely and efficiently solve challenging autonomous driving tasks in complex mixed traffic environments. It significantly outperforms model-free RL and BC in terms of learning efficiency and performance indices. In future research, we will extend the proposed method to address more complex static road topology scenarios encountered in urban conditions.

\section*{Acknowledgments}
We would like to thank the National Key R\&D Program of China under Grant No2022YFB2502900, National Natural Science Foundation of China (Grant Number: U23B2061), the Fundamental Research Funds for the Central Universities of China, Xiaomi Young Talent Program, and thanks the reviewers for the valuable suggestions

\bibliographystyle{IEEEtran}
\bibliography{document}

\vspace{-3.5em}
\begin{IEEEbiography}[{\includegraphics[width=1in,height=1.25in,clip,keepaspectratio]{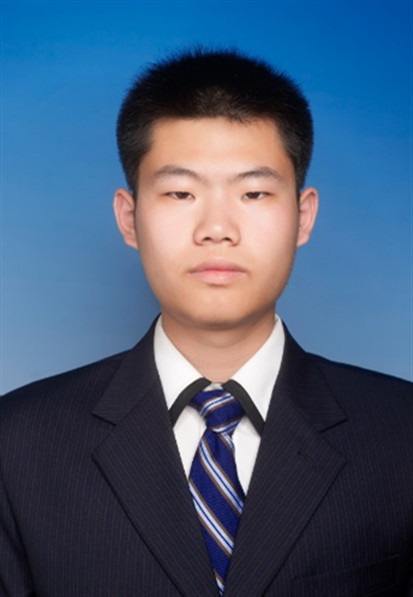}}]{Shuo Yang} received the B.S. and M.S degree (cum laude) from the College of Automotive Engineering, Jilin University, Changchun, China, in 2017. He is currently pursuing the Ph.D. degree in School of Automotive Studies, Tongji University, Shanghai. His research interests include reinforcement learning,  autonomous vehicle, intelligent transportation system and vehicle dynamics.
\end{IEEEbiography}
\vspace{-3.5em}

\begin{IEEEbiography}[{\includegraphics[width=1in,height=1.25in,clip,keepaspectratio]{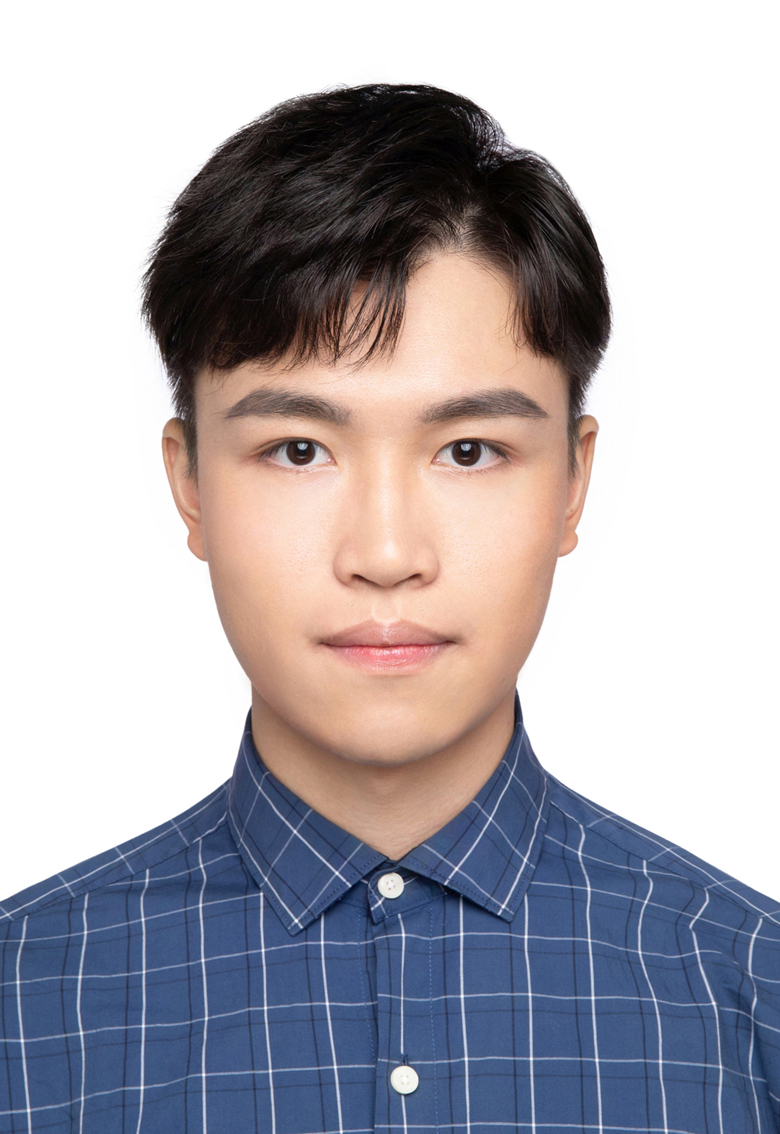}}]{Caojun Wang} received the B.S. degree (cum laude) from the School of Mechatronic Engineering And Automation, Shanghai University, Shanghai, China, in 2023. He is currently pursuing the Ph.D degree in School of Automotive Studies, Tongji University, Shanghai, China. His research interests include reinforcement learning and autonomous vehicle.
\end{IEEEbiography}
\vspace{-3.5em}

\begin{IEEEbiography}[{\includegraphics[width=1in,height=1.25in,clip,keepaspectratio]{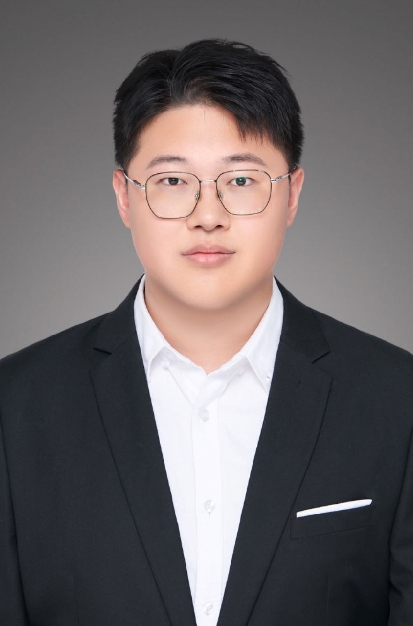}}]{Zhenyu Ma} received the B.S. degree in College of Automotive Engineering, Jilin University, Jilin, China, in 2022. He is currently working toward the M.S. degree with the School of Automotive Studies, Tongji University, Shanghai, China. His research interests include machine learning, reinforcement learning, decision control, intelligent transportation systems, and autonomous vehicle.
\end{IEEEbiography}
\vspace{-4.5em}

\begin{IEEEbiography}[{\includegraphics[width=1in,height=1.25in,clip,keepaspectratio]{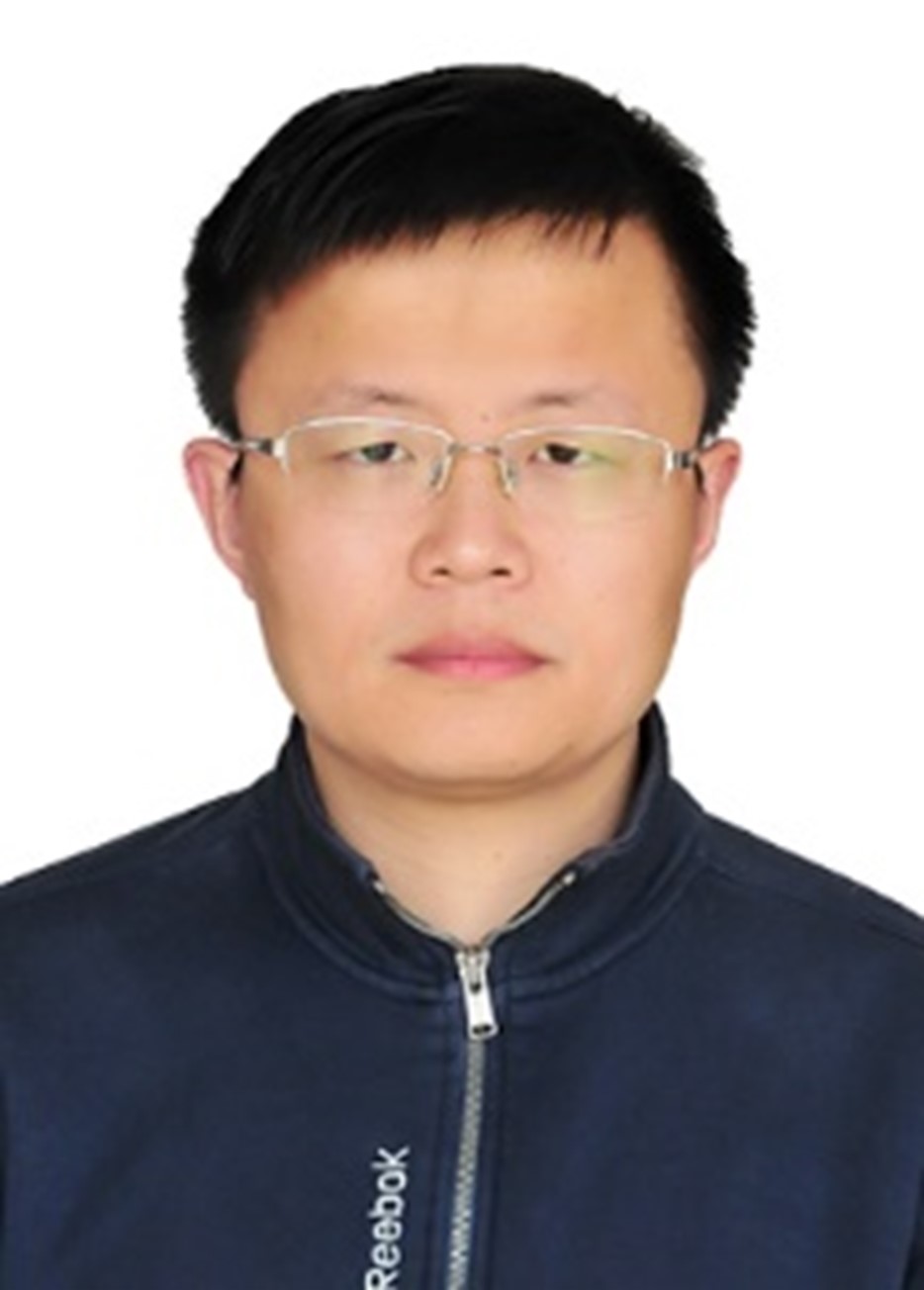}}]{Yanjun Huang}
is a Professor at School of Automotive studies, Tongji University. He received his PhD Degree in 2016 from the Department of Mechanical and Mechatronics Engineering at University of Waterloo. His research interest is mainly on autonomous driving and artificial intelligence in terms of decision-making and planning, motion control, human-machine cooperative driving. He has published several books, over 80 papers in journals and conference; He is the recipient of IEEE Vehicular Technology Society 2019 Best Land Transportation Paper Award. He is serving as AE of IEEE/TITS, P I MECH ENG D-JAUT, IET/ITS, SAE/IJCV, Springer Book series of connected and autonomous vehicle, etc.
\end{IEEEbiography}
\vspace{-3.5em}

\begin{IEEEbiography}[{\includegraphics[width=1in,height=1.25in,clip,keepaspectratio]{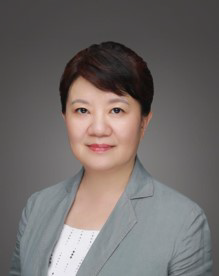}}]{Hong Chen }
(M'02-SM'12-F'22) received the B.S. and M.S. degrees in process control from Zhejiang University, China, in 1983 and 1986, respectively, and the Ph.D. degree in system dynamics and control engineering from the University of Stuttgart, Germany, in 1997. In 1986, she joined Jilin University of Technology, China. From 1993 to 1997, she was a Wissenschaftlicher Mitarbeiter with the Institut fuer Systemdynamik und Regelungstechnik, University of Stuttgart. Since 1999, she has been a professor at Jilin University and hereafter a Tang Aoqing professor. Recently, she joined Tongji University as a distinguished professor. Her current research interests include model predictive control, nonlinear control, artificial intelligence and applications in mechatronic systems e.g. automotive systems.
\end{IEEEbiography}

\end{document}